# Cross-domain attribute representation based on convolutional neural network[*]

Guohui Zhang[1], Gaoyuan Liang[2], Fang Su[3], Fanxin Qu[4], and Jing-Yan Wang[5]

[1]Huawei Technologies Co., Ltd., Shanghai, China
guohuizhang354@outlook.com
[2]Jiangsu University of Technology, Jiangsu 213001, China
gaoyuanliang@outlook.com
[3]Shaanxi University of Science & Technology, Xi'an, China
[4]Northwestern Polytechnical University, Xi'an, China
[5]Provincial Key Laboratory for Computer Information Processing Technology,
Soochow University, Suzhou 215006, China
jimjywang@gmail.com

**Abstract.** In the problem of domain transfer learning, we learn a model for the prediction in a target domain from the data of both some source domains and the target domain, where the target domain is in lack of labels while the source domain has sufficient labels. Besides the instances of the data, recently the attributes of data shared across domains are also explored and proven to be very helpful to leverage the information of different domains. In this paper, we propose a novel learning framework for domain-transfer learning based on both instances and attributes. We proposed to embed the attributes of different domains by a shared convolutional neural network (CNN), learn a domain-independent CNN model to represent the information shared by different domains by matching across domains, and a domain-specific CNN model to represent the information of each domain. The concatenation of the three CNN model outputs is used to predict the class label. An iterative algorithm based on gradient descent method is developed to learn the parameters of the model. The experiments over benchmark datasets show the advantage of the proposed model.

**Keywords:** Convolutional Neural Network, Domain-Transfer Learning, Attribute Embedding.

## 1 Introduction

In the machine learning problems, domain transfer learning has recently attracted much attention [23,26]. Transfer learning refers to the learning problem of a predictive model for a target domain, by leveraging the data from both the target domain and one or more auxiliary domains. One shortage of traditional transfer learning methods is that the attributes of the data are not used by the classification model. But the attributes of the data actually has the nature of stability across the domains. Thus using the attributes of the data is critical for the transfer learning [21,19]. Peng et al. [19] proposed to represent the attribute vectors of each data point by using an attribute dictionary. Each data point is reconstructed by the elements in the dictionary, and the reconstruction coefficients are used as the new representation of the attributes. The attribute vector of a data point is mapped to the new representation vector by a linear transformation matrix so that the new representation vector is linked to the attribute vector. To leverage the auxiliary and the target domains, the same attribute representation method is applied to both auxiliary and target domains. The learning process is regularized by the class-intra similarity in the auxiliary domains, and by the neighborhood in the target domain. Su et al. [21] proposed a low-rank attribute embedding method for the problem of person re-identification of multiple cameras. The proposed method tries to solve the problem of multiple cameras based person re-identification as a multi-task learning problem. The proposed method uses both the low-level features with mid-level attributes as the input of the identification model. The embedding of attributes maps the attributes to a continuous space to explore the correlative relationship between each pair of attributes and also recovers the missing attributes. Both these two methods of attribute representation are based on the linear transformation. However, a simple linear function may be insufficient to represent the attributes effectively.

[*] The study was supported by Provincial Key Laboratory for Computer Information Processing Technology, Soochow University, China (Grant No. KJS1324).

In this paper, we propose a novel attribute embedding method for attributes for the problem of domain transfer learning. The embedding of attributes is based on convolutional neural network (CNN) model [25,12,13]. The convolutional output of the input data is further mapped to the attribute vector. In this way, the attribute embedding vector not only represents the attributes of a data point but also contains the pattern of the input data constructed by the CNN model, which has been proven to be a powerful representation model. To construct the classification model for each domain, we also learn a domain independent convolutional representation and a domain-specific convolutional representation. The domain-independent convolutional representation maps the data of different domains to a shared data space to capture the patterns shared over all the domain. The domain-specific convolutional representation is used to represent the patterns specifically contained by each domain. The classification model of each domain is based on the three types of convolutional representations, i.e., attribute embedding, domain-independent and domain-specific representations. To learn the parameters of the models, we propose to minimize the mapping errors of the attributes, the classification errors across different domains, the mismatching of different domains in the domain-independent representation space, and the dissimilarity between the neighboring data points in the target domain. The joint minimization problem is solved by an alternate optimization strategy and the gradient descent algorithm.

## 2 Method

In the problem setting of cross-domain learning, we assume we have T domains. The first T − 1 domain are the auxiliary domains, while the T-th domain is the target domain. The problem is to learn an effective model for the classification of the target domain. The input data sets of the T domains are denoted as $X_t|_{t=1}^T$, where $X_t = \{(X_t^i, a_i^t, y_i^t)\}$ is the data set of the t-th domain. $X_i^t = [x_{i1}^t, \cdots, x_{i|X_i^t|}^t] \in R^{d \times |X_i^t|}$ is the input matrix of the i-th data point of the t-th domain, and each column of the matrix is a feature vector of a instance, and $a_i^t \in \{1,0\}^{|a|}$ is its binary attribute vector, and $y_i^t \in \{1,0\}^{|y|}$ is its class label vector.

We propose to embed the attributes of each input data point to a vector and use the convolutional representation of the input data as the embedding vector. Given the the input matrix of a data point, X, we represent it by a CNN model composed of a convolutional layer, a activation layer, and max-pooling layer, denoted as $f_a(X)$. Since this convolutional representation of X is used as its attribute vector embedding, we propose to map it to the attribute vector a by a linear mapping function,

$$f_a(X) \leftarrow \Theta^T a \quad (1)$$

where $\Theta \in R^{|a| \times m}$ is the mapping matrix. To reduce the mapping errors, we proposed to minimize the Frobenius norm distance between the convolutional representations and the mapping results for all the data points of all domains,

$$\min_{\Theta, f_a} \sum_{t=1}^{T} \left( \sum_{i=1}^{n_t} \left\| f_a(X_i^t - \Theta a_i^t) \right\|_F^2 \right) \quad (2)$$

To predict the class labels for the data points in multiple domains, we proposed the data of each domain to represent the data into a domain-independent convolutional representation and a domain-specific convolutional representation. The domain-independent convolutional representation function is shared across all the domains. It tries to extract features relevant to the class labels, but independent of the specific domains. The domain-independent convolutional recreation function is also based on a CNN model, denoted $f_0(X)$. Since the $f_0(X)$ outputs are domain-independent, we hope the representations of data points from different domains can be similar to each other. To this end, we impose that the distribution of the base representations of different domains is of the same. We use the mean vector of the representations of each domain as the presentation of the distribution of the domain. For the t-th domain, the mean vector is given as $\frac{1}{n_t} \sum_{i=1}^{n_t} f_0(X_i^t)$. To reduce the mismatch among the domains, we proposed to minimize the Frobenius norm distances between the mean vectors of each pair of domains,

$$\min_{f_0} \sum_{t,t'=1,t<t'}^{T} \left\| \frac{1}{n_t} \sum_{i=1}^{n_t} f_0(X_i^t) - \frac{1}{n_{t'}} \sum_{i=1}^{n_{t'}} f_0(X_i^{t'}) \right\|_F^2 \quad (3)$$

To predict the class labels for the data points of different domains, we also consider the representation of the data points according to the domains. This is the domain-specific representation. The representation is also based on CNN models, and the CNN model of the t-th domain of a data point X is denoted as $f_t(X)$.

To estimate the class label from a data point of the t-th domain, we concatenate both the domain-independent and domain-specific convolutional representations of the input data, $f_0(X)$ and $f_t(X)$, and also the attribute embedding of the data, $f_a(X)$. They are concatenated to a longer vector,

$$f(X) = \begin{bmatrix} f_0(X) \\ f_t(X) \\ f_a(X) \end{bmatrix} \in R^{m_0+m_t+m_a} \quad (4)$$

and the longer vector is transformed to a $|y|$-dimensional vector of scores of classification by a matrix $U = \begin{bmatrix} U_0 \\ U_t \\ U_a \end{bmatrix} \in R^{(m_0+m_t+m_a) \times |y|}$ in a classification function,

$h_t(X) = U^T f(X) = U_0^T f_0(X) + U_t^T f_t(X) + U_a^T f_a(X)$, $t = 1, \cdots, T$, where $U_0$, $U_t$, and $U_a$ are the transformation matrices for the domain-independent representation, domain-specific representation, and the attribute embedding.

To learn the parameters of the model, we propose the following minimization,

$$\min_{U_t, W_t|_{t=0}^T, U_a, W_a, \Theta} \left\{ o(U_t, W_t|_{t=0}^T, U_a, W_a, \Theta) = \sum_{t=1}^{T-1} \left( \sum_{i=1}^{n_t} \left\| y_i^t - h_t(X_i^t) \right\|_F^2 \right) \right.$$
$$+ \sum_{i=1}^{l_T} \left\| y_i^T - h_t(X_i^T) \right\|_F^2 + C_1 \sum_{t=1}^{T} \left( \sum_{i=1}^{n_t} \left\| f_a(X_i^t) - \Theta a_i^t \right\|_F^2 \right)$$
$$+ C_2 \sum_{t,t'=1,t<t'}^{T} \left\| \frac{1}{n_t} \sum_{i=1}^{n_t} f_0(X_i^t) - \frac{1}{n_{t'}} f_0(X_i^{t'}) \right\|_F^2 \quad (5)$$
$$\left. + C_3 \sum_{i,i'=1}^{n_T} M_{ii'} \left\| f(x_i^T) - f(X_{i'}^T) \right\|_F^2 \right\}$$

We explain the objective terms of the objective function o as follows,

— The classification function is used to predict the class labels, thus we propose to reduce the prediction errors measured by the Frobenius norm distance between the class label vectors and the outputs of ht(X) for the data pints with available label vectors. The first two terms are the classificaiton error terms.
— The third term is attribute mapping error term of (2).
— The fourth term is the cross-domain matching term of the domain-independent CNN model of (3).
— For the unlabeled data points in the target domain, we also regularize them by imposing their representations to be constant with the labeled data points in the neighborhood, so that the supervision information can also be propagated to them. To this end, we hope for any neighboring two data points in the target domain, their overall representation vectors are close to each other. To this end, in the last term, we minimize the Frobenius norm distance between the representations of neighboring data points in the target domain, where $M_{ii'} = 1$ if $X_i$ and $X_{i'}$ are a neighbor to each other and 0 otherwise.

$C_k, k = 1, \cdots, 3$ are the tradeoff weights of different regularization terms. To solve this problem, we proposed to use the sub-gradient descent algorithm. The filters of the CNN models, the mapping matrix, and the transformation matrices are updated according to the direction of gradient function of the objective,

$$\Phi \leftarrow \Phi - \tau \nabla_\Phi o, \qquad (6)$$

$$\text{where } \Phi = \{f_a, f_0, f_1, \cdots, f_T, \Theta, U_0, U_a, U_1, \cdots, U_T\},$$

$\tau$ is the descent step, and $\nabla_\Phi o$ is the sub-gradient of o regarding $\Phi$. In an iterative, the parameters of $\Phi$ are updated alternately until a maximum iteration number is reached or converge is achieved.

## 3 Experiment

In this section, we evaluate the proposed method over several domain-transfer problems.

### 3.1 Data sets and experimental setting

In the experiments, we use three datasets as follows. CUHK03 data set was developed for the problem of person re-identification problems [15]. It contains 13,164 images of 1,360 persons. For each image, we annotate it by 108 attributes, including gender (male/female), wearing long hair, etc. The images are captured by six different cameras. The problem of person re-identification is to train a classifier over the images of some cameras, and then use the classifier to identify an image captured from other cameras. We treat each camera as a domain, and we use each domain as a target domain in turn. Bankrupt prediction data contains the stock price wave data of 3 years of 374 companies of three different countries, China, USA, and UK. We collected this data for the problem of prediction of company bankrupt. Each company is also labeled by a list of business type attributes. Each company is treated as a data points, presented by a set of short-term frames, and a list of binary attributes of business types. Moreover, each country is treated as domain. The prediction problem of this data set is to predict if a given company will be in bankrupt within the future 3 years. Spam email data set is for the spam email detection competition of the ECML/PKDD Discovery Challenge 2006 [3]. It contains texts of emails of 15 email users, and for each user, there are 400 emails. Among the 400 emails of each user, half of them are spam emails, while the remaining half are non-spam emails. Each email text is composed of a set of words. Moreover, we also apply a topic classifier and a sentiment classifier to each email text to extract attributes of the text and use the extracted attributes as additional information. Each user is treated as a domain, and we also use each user as a target domain in turn. In our experiments, given a data set of several domains, we treat each domain as a target domain in turn, while treating the other domains as the auxiliary domains to help train the model. The data points in a target domain are further split into a training set and a test set with equal sizes randomly. Meanwhile, for the training set of the target domain, we further split it into equal-sized subsets. One subset is used as a labeled set, and the other set is used as an unlabeled set. We train the model over the data points of the auxiliary domains and the training set of the target domain and then test it over the test set of the target domain. The classification rate over the test set is used as the performance measure. The average classification rate over different target domains is reported and compared.

### 3.2  Results

In the experiments, we first compare the proposed domain transfer convolutional attribute embedding (DTCAE) algorithm to some state-of-the-art domain-transfer attribute representation methods, and then study the properties of the proposed algorithm experimentally.

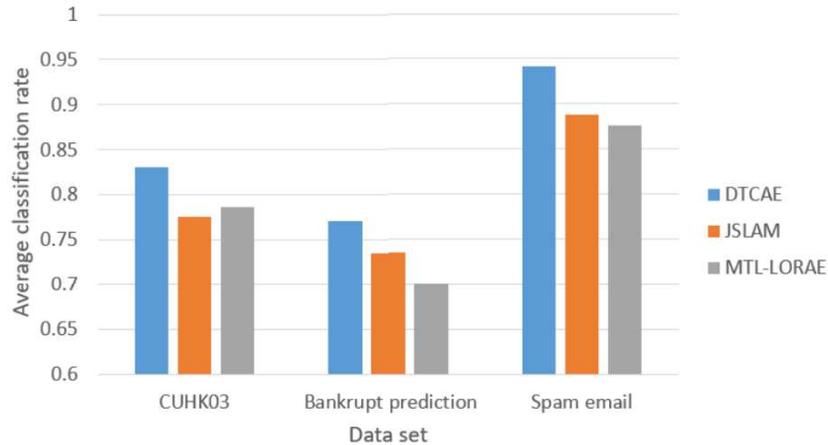

**Fig.1**. Comparison results over the benchmark data sets.

Attribute embedding for domain-transfer learning problem is a new topic and there are only two existing methods. In the experiment, we compare the proposed algorithm against the two existing methods, which are the Joint Semantic and Latent Attribute Modelling (JSLAM) method proposed by Peng et al. [19], and the Multi-Task Learning with Low-Rank Attribute Embedding (MTL-LORAE) method proposed by Su et al. [21]. The comparison results over the three benchmark data sets are shown in Figure 1. According to the reported average accuracies over the benchmark datasets, our algorithm DTCAE achieves the best performance over all the three datasets. For example, over the CUHK03 data set, the DTCAE is the only compared method which has an average accuracy higher than 0.800. Meanwhile, over the spam email dataset, only DTCAE obtains an average accuracy higher than 0.900.

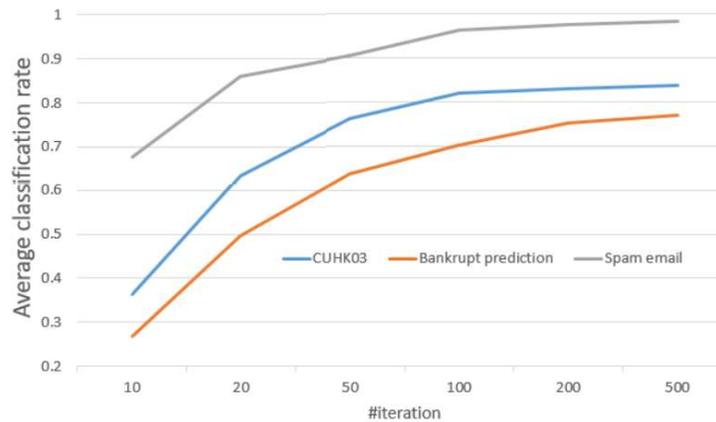

**Fig.2**. Convergence curves over the benchmark data sets.

Since the proposed algorithm DTCAE is an iterative algorithm. The variables are updated alternately. We are also interested in the convergence of the algorithm. Thus we plot the average classification rates with a different number of iterations. The curves over the three benchmark data sets are plotted in Figure 2. According to the curves of Figure 2, when more iterations are used to update the variables of the model, the average classification rates increase stably. This is not surprising because a larger number of iterations reaches a smaller objective function. This verifies the effectiveness of the proposed model and its corresponding objective function. Moreover, we also observe that when the iteration number is larger than 100, the change of the performance is very small. This means that the algorithm converges and no more iteration is needed to improve the performance.

## 4      Conclusion

In this paper, we propose a novel model for the problem of cross-domain learning problem with attribute data. We use a CNN model to map the input data to its attributes.

Moreover, a domain-independent and domain-specific CNN model are also used to represent the data input itself. The attribute embedding, the domain-independent, and domain-specific representations are concatenated as the new representation of the data points, and we further a linear layer to map the new representation to the class labels. Moreover, we also impose the domain-independent representations of data points of different domains to be in a common distribution, and the neighboring data points of target domain to be similar to each other. We model the learning problem as a minimization problem and solve it by an iterative algorithm. The experiments on three benchmark data sets show its advantages. In the future, we plan to apply the proposed method other applications, such as biometrics [20,27], network analysis [24], human pose estimation [11,10,9], mobile computing [16,8], mathematics [17,2,1,18], etc. The similar approach can also be adopted in other related fields such as systems [6,7,14], and system security [4,5,22].